\documentclass{article}

\PassOptionsToPackage{numbers, compress, sort}{natbib}
\usepackage[preprint]{neurips_2026}

\usepackage[utf8]{inputenc} 
\usepackage[T1]{fontenc}    
\usepackage{hyperref}       
\usepackage{url}            
\usepackage{booktabs}       
\usepackage{amsfonts}       
\usepackage{nicefrac}       
\usepackage{microtype}      
\usepackage{xcolor}         
\usepackage{amsmath}
\usepackage{enumitem}

\usepackage{booktabs}
\usepackage{multirow}
\usepackage{makecell}
\usepackage{graphicx}
\usepackage{amssymb}
\usepackage{pdflscape}
\usepackage[table]{xcolor}
\usepackage{tabularx}
\usepackage{tcolorbox}
\usepackage{wrapfig}
\usepackage{dsfont}
\usepackage{wrapfig}
\usepackage[nameinlink,capitalise,noabbrev]{cleveref}
\usepackage[ruled,linesnumbered,vlined]{algorithm2e}
\crefname{figure}{Fig.}{Figs.}
\Crefname{figure}{Fig.}{Figs.}
\crefname{table}{Tab.}{Tabs.}
\Crefname{table}{Tab.}{Tabs.}
\crefname{section}{Sec.}{Secs.}
\Crefname{section}{Sec.}{Secs.}
\crefname{equation}{Eq.}{Eqs.}
\Crefname{equation}{Eq.}{Eqs.}

\title{CAIRN: Cross-Room 3D Scene Understanding with Topology-Aware Large Multimodal Models}

\author{%
  He Liang\textsuperscript{1}\thanks{Part of this work was done during an internship at Microsoft Research.} \quad
  Chenyang Ma\textsuperscript{1} \quad
  Yiming Zhang\textsuperscript{3} \quad
  Sangyun Shin\textsuperscript{1} \\
  Andrew Markham\textsuperscript{1} \quad
  Niki Trigoni\textsuperscript{1} \quad
  Yuhang He\textsuperscript{2}\thanks{Corresponding author} \\
  \\
  \textsuperscript{1}University of Oxford \quad
  \textsuperscript{2}Microsoft Research\quad
  \textsuperscript{3}Simon Fraser University \\
  \texttt{\{he.liang, chenyang.ma, sangyun.shin, andrew.markham, niki.trigoni\}@cs.ox.ac.uk} \\
  \texttt{yuhanghe@microsoft.com} \quad
  \texttt{yza440@sfu.ca}
}

\begin{document}
\maketitle
\begingroup
\renewcommand{\thefootnote}{}
\footnotetext{\textbf{Project Page:}
\href{https://oceansdepp.github.io/cairn_web/}
{\nolinkurl{https://oceansdepp.github.io/cairn_web/}}}
\endgroup
\begin{abstract}
Existing 3D scene-grounded Large Language Models (3D-LLMs) focus on answering questions grounded in simplified single-room 3D scenes, lacking the ability to reason over real-world household environments containing multiple interconnected rooms and diverse object categories. We introduce \textbf{CAIRN}, a topology-aware 3D-LLM for multi-room 3D scene understanding. CAIRN aligns transformer attention with scene hierarchy, giving the model explicit awareness of object-level relations and room-level connectivity. It enriches object tokens with room-local relational context via a graph neural network, introduces learned room tokens for room-level abstraction, and applies a hierarchical attention mask with geometric bias to route information according to scene topology. CAIRN is developed on \textbf{CAIRN-MR}, a benchmark we introduce on HM3D for multi-room 3D scene understanding, covering grounding, captioning, and four question-answering tasks that progressively evaluate from intra-room perception to cross-room reasoning. Experiments show that CAIRN outperforms prior 3D-LLMs by a large margin across all CAIRN-MR tasks while remaining competitive on five single-room benchmarks.
\end{abstract}
\section{Introduction}
\label{sec:intro}

A capable home assistant agent should recall where an object was last seen in the kitchen and retrieve it from another room, or answer queries that relate objects across spaces (e.g., comparing furniture arrangements across bedrooms). This requires understanding the full 3D scene structure of a household and performing consistent reasoning over the entire multi-room environment. 

There has been a rich history of prior work on developing agents that answer questions grounded in 3D scenes, using large-language models (LLMs) augmented with structured 3D scene representations (i.e., 3D-LLMs) as the reasoning backbone~\cite{li2025videochat, liu2023improved, llava, lisa, ferret, zhao2023bubogpt}. However, these approaches largely restrict 3D scene understanding and spatial reasoning to a single room~\cite{chatscene, chat3dv2, 3dgraphllm, grounded3dllm, leo, ll3da, qi2025gpt4scene}, representing scenes as flat sequences of object tokens without modeling higher-level room structure. In this work, we aim to move beyond the single-room paradigm and develop 3D-LLMs that can understand full household settings~\cite{puig2023habitat}, where multiple interconnected rooms contain hundreds of objects. This unlocks more challenging tasks such as grounding across spatial contexts, tracing relationships between rooms, and reasoning over a scene’s topological structure. 

To enable the study of multi-room 3D scene understanding, we first introduce \textbf{CAIRN-MR}, a benchmark for structured multi-room 3D scene understanding built on HM3D~\cite{ramakrishnan2021habitat}, as existing 3D scene-language datasets~\cite{scanrefer, multi3drefer, scan2cap, scanqa, sqa3d} are mostly limited to single-room scans. Unlike prior benchmarks where reasoning is confined to a single room with a limited candidate set, CAIRN-MR shifts the problem to full-scene reasoning over multi-room environments, where models must search over larger spaces, handle similar object configurations, and resolve ambiguities across spatially separated regions. Crucially, the target region is not explicitly specified, requiring models to first localize the relevant room(s) from implicit object-level spatial constraints before performing downstream reasoning. Motivated by these challenges, CAIRN-MR preserves grounding and captioning tasks from prior benchmarks for backward compatibility, while further introducing four question answering tasks that progressively evaluate multi-room reasoning under increasingly challenging settings, ranging from room-grounded perception to cross-room comparison. 



Developing 3D-LLMs for multi-room 3D scene understanding introduces challenges beyond the single-room setting. First, existing 3D-LLMs represent a scene as a flat set of object tokens with self-attention over all object pairs~\cite{chatscene, leo, ll3da, 3dgraphllm, yu2025inst3d}. While this design is adequate for single-room scenes, it is poorly suited to multi-room environments, where interactions are primarily intra-room, and cross-room dependencies are sparse and structured by room connectivity. Second, multi-room scenes contain geometric and topological priors, such as relative object poses, distances, and room adjacency, that should be explicitly modeled, but are not exploited by standard self-attention. To address these challenges, we propose \textbf{CAIRN}, a topology-aware 3D-LLM that explicitly models the hierarchical structure and topology of multi-room 3D scenes. CAIRN first encodes each scene as a hierarchical 3D scene graph, capturing both object-level relations and room-level connectivity. After tokenizing this graph into scene tokens that preserve both object- and room-level information, CAIRN uses masked attention to handle the sparse and structured interactions in multi-room scenes by constraining information flow according to scene topology. It further incorporates geometric bias into attention to encode explicit spatial priors, such as relative poses, distances, and room adjacency. In summary, our contributions are as follows:

\begin{itemize}[leftmargin=*]
\renewcommand\labelitemi{\scalebox{0.75}{$\bullet$}}
\item We present the first study of 3D-LLMs for multi-room scene understanding. We introduce \textbf{CAIRN-MR}, a benchmark for multi-room 3D scene understanding that covers grounding, captioning, and four question-answering task types spanning intra-room perception to cross-room comparison.
\item We propose \textbf{CAIRN}, a topology-aware 3D-LLM for multi-room understanding that uses masked attention to align information flow with scene topology and route information across rooms, and uses geometric bias to inject explicit spatial priors into attention.
\item Extensive experiments show that CAIRN outperforms prior methods by a large margin on CAIRN-MR across all tasks, while remaining effective on five single-room 3D scene understanding benchmarks without compromising fine-grained object-level understanding. We validate the quality of CAIRN-MR through human calibration and statistical baselines.
\end{itemize}

\begin{figure}[t]
    \centering
    \includegraphics[width=\linewidth]{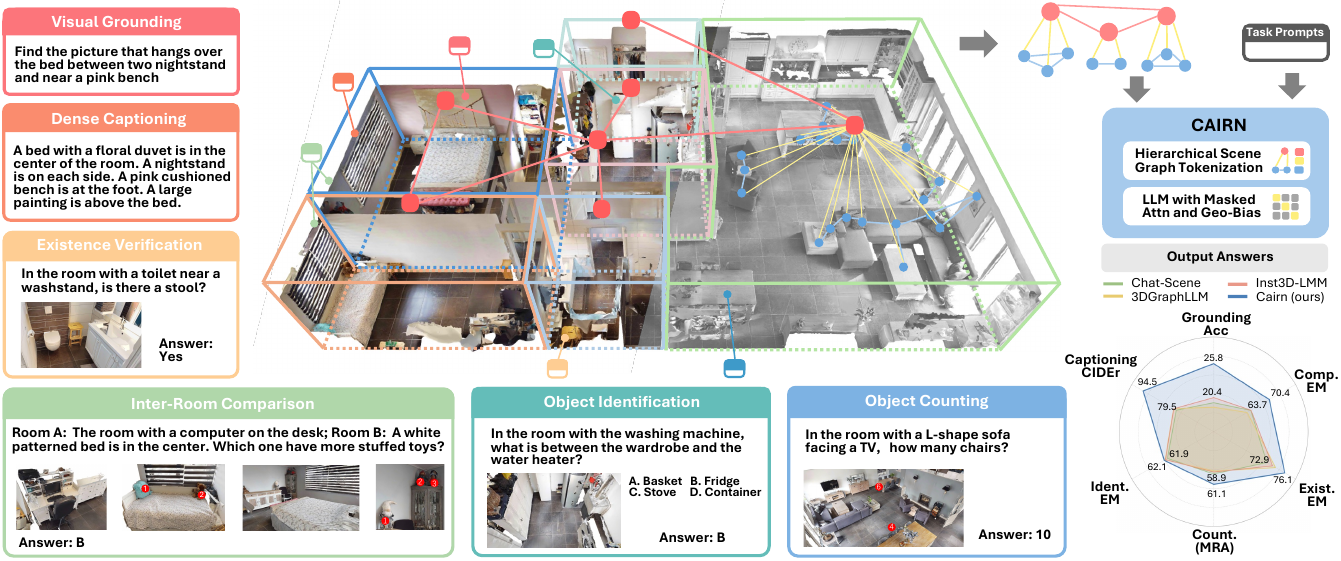}
    \vspace{-6mm}
    \caption{\textbf{3D-LLMs for multi-room scene understanding.} We introduce \textbf{CAIRN}, a topology-aware 3D-LLM for multi-room scene understanding, along with \textbf{CAIRN-MR}, a multi-room 3D scene understanding benchmark with diverse tasks. By representing each scene as a hierarchical scene graph and aligning attention with scene topology through structured masking and geometric bias, CAIRN achieves substantial gains over prior 3D-LLMs.}
    \label{fig:teaser}
    \vspace{-1.2em}
\end{figure}



\section{Related Work}

\noindent\textbf{3D-LLMs and Benchmarks.} 
Most 3D-LLMs are developed on single-room benchmarks from ScanNet~\cite{scannet, scanrefer, scanqa, scan2cap, sqa3d, multi3drefer}, and thus represent scenes as flat sets of object tokens with dense self-attention. Some adopt object-level tokenization with unique identifiers~\cite{chat3dv2, chatscene, grounded3dllm}, while others improve scene reasoning through training strategies~\cite{leo, ll3da, scenellm}, data~\cite{kang2024robin3d, qi2025gpt4scene, 3urllm}, or input design~\cite{3dllm, lscenellm, lu2025kitchenvla, pts3dllm, ma2024spatialpin, descrip3d}. Some incorporate spatial relations via graph networks~\cite{3dgraphllm, yu2025inst3d}. However, none models rooms as semantic units or routes information by scene topology. While some recent efforts extend to larger scenes~\cite{lscenellm}, they often construct queries with globally unique targets, effectively reducing the ambiguity inherent in multi-room environments. These settings primarily evaluate large-scale perception rather than structured multi-room reasoning, and do not capture key structural properties such as implicit room hierarchy, recurring object categories across rooms, and layout ambiguity. To our knowledge, CAIRN is the first 3D-LLM to introduce room-level abstraction and topology-aware attention, together with a multi-room benchmark for evaluation.

\noindent\textbf{Multi-Room Embodied Tasks.} Simulators that provide multi-room household environments~\cite{puig2018virtualhome, khanna2023hssd, puig2023habitat} enable challenging multi-room tasks such as human-robot collaboration~\cite{puig2020watch, chang2024partnr}, long-horizon robot task planning~\cite{rana2023sayplan, ma2025coopera, ma2026cyclevla, honerkamp2024language}, and navigation~\cite{wu2024camon, yenamandra2023homerobot}. These works typically rely on ground-truth maps or pre-built scene graphs. Complementary efforts construct hierarchical 3D scene graphs from raw data~\cite{hughes2022hydra, werby2024hierarchical, gu2024conceptgraphs}, but do not study language-grounded reasoning over such representations. Our work bridges these two directions by building hierarchical representations from raw 3D data and enabling language-grounded multi-room scene understanding over them.

\noindent\textbf{Attention with Structural Priors.}
Standard self-attention models dense pairwise interactions without structural awareness. Many works revise this by injecting domain-specific priors. Sparse attention methods~\cite{beltagy2020longformer, zaheer2020big, child2019generating} restrict attention to local windows or global tokens for efficiency in long sequences. Others inject structural priors as additive biases, such as ALiBi~\cite{press2021train}, which penalizes attention by linear distance, and Graphormer-style models~\cite{ying2021transformers, liu2024gradformer, chen2022structure}, which encode shortest-path distances and node centrality. We instead design topology-aware attention for multi-room scenes, combining masks that constrain information flow with geometric biases that encode spatial relations.
\section{CAIRN-MR: Multi-Room 3D Scene Understanding Benchmark}
\label{sec:task}
\begin{figure}[t]
    \centering
    \vspace{-6mm}
    \includegraphics[width=\linewidth]{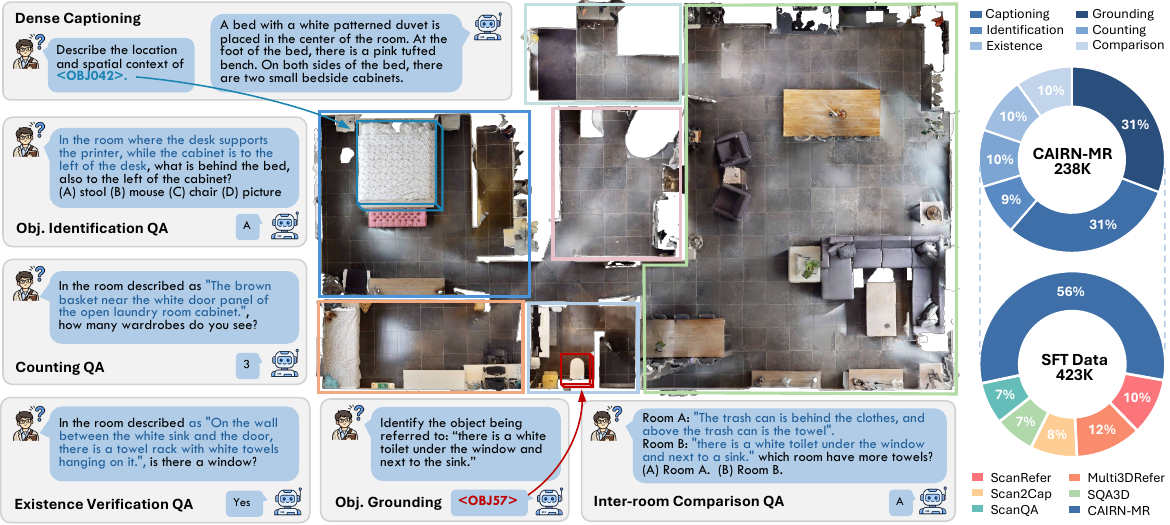}
    \caption{\textbf{CAIRN-MR benchmark for multi-room 3D scene understanding.} The benchmark includes grounding, captioning, and question answering tasks covering room localization, intra-room reasoning, and cross-room comparison. Colored boxes denote parsed room regions. Right: benchmark scale and task composition.}
    \label{fig:dataset}
    \vspace{-1.2em}
\end{figure}

Most existing 3D scene understanding benchmarks confine reasoning to single-room scenes, yet real household environments contain multiple interconnected rooms with distinct spatial structure. To advance 3D-LLMs toward structured multi-room scene understanding, we introduce \textbf{CAIRN-MR}, the first benchmark covering six cross-room reasoning tasks (~\Cref{fig:dataset}). These tasks capture challenges overlooked in single-room settings: objects of the same category may appear across different rooms, requiring disambiguation over spatially separated regions; reasoning must span both object-level details and room-level structure; and scene information grows substantially with the number of rooms, demanding more scalable understanding methods.


\subsection{Multi-Room Reasoning and Task Design}
\label{sec:tasks}
CAIRN-MR presents three progressively more challenging forms of multi-room reasoning. It remains compatible with standard grounding and captioning tasks, while introducing question answering tasks that directly test room hierarchy, implicit room localization, and cross-room ambiguity.

\noindent\textbf{1) Object Visual Grounding.} Given a referring expression, the model localizes the target object among all objects in the scene. Unlike prior settings where the search is confined to a single region, disambiguation here must span the entire scene, and recurring object categories with similar layouts across rooms make multiple candidates plausible matches.

\noindent\textbf{2) Dense Captioning.} Given a target object, the model generates a referring expression that uniquely identifies it. Unlike single-room settings, the expression must distinguish the target from all same-category instances across the entire scene, not just those within one room. This requires the model to reason about which attributes or spatial relations remain discriminative at scene scale.

\noindent\textbf{3) Object Identification.} Given a spatial description that implicitly specifies a room, together with an additional cue about a target object inside it, the model predicts the target category from four candidates. This requires two-stage reasoning: the model must first localize the room from the spatial description, and then identify the target object within that room. Errors in room localization therefore directly affect the final prediction.

\noindent\textbf{4) Counting Question Answering.} The model counts all instances of a queried category within the room implicitly specified by the spatial description. The main challenge is restricting the count to the correct room: same-category instances often appear in multiple rooms, so accurate counting depends on correct room localization.

\noindent\textbf{5) Existence Verification Question Answering.} The model determines whether a queried category is present in the room specified by the spatial description. Negative cases are particularly challenging, because the queried category may still appear elsewhere in the scene. To answer correctly, the model must reason over the localized room rather than relying on scene-level presence.

\noindent\textbf{6) Inter-room Comparison Question Answering.} Given two spatial descriptions that specify two different rooms, the model determines which room contains more instances of a queried category. This task introduces explicit cross-room reasoning: both rooms must be localized and counted separately, and an error in either room is sufficient to invalidate the comparison.

\subsection{Benchmark Construction and Statistics}
Tasks in CAIRN-MR require the model to infer the relevant room from object-level spatial descriptions rather than predefined room labels. This places a strong requirement on the referring expressions: they must be spatially discriminative enough to uniquely identify objects across the full scene and structurally rich enough to implicitly specify which room is being referenced. To achieve this, we construct the benchmark in three stages: scene assembly, expression generation, and task instantiation. We build on HM3D~\cite{ramakrishnan2021habitat} residential environments and the object-level relational annotations from SceneVerse~\cite{jia2024sceneverse}. In the first stage, we assemble connected multi-room layouts from proximate rooms in each HM3D building, ensuring mutual reachability. In the second stage, we construct referring expressions from SceneVerse spatial relations in three forms--multi-reference, multi-hop chains, and branched chains--and retain those that are sufficiently informative to uniquely identify a single object across all rooms. A subset of expressions is further augmented with VLM-derived appearance modifiers (e.g., chair $\rightarrow$ wooden chair) to increase linguistic diversity. In the third stage, the validated expressions are used to instantiate all six tasks, including grounding, captioning, and four QA types via LLM rewriting and rule-based templates. Full details are provided in Appendix~\ref{app:bench_construction}.

The resulting benchmark contains 673 scenes, with 479 for training and 194 for validation. Splits are partitioned at the building level with no scene overlap. On average, each scene contains 4.5 rooms and approximately 115 object instances. As summarized in~\Cref{fig:dataset}, CAIRN-MR provides 238K task annotations spanning grounding, captioning, and question answering.

\section{CAIRN: Hierarchical Topology-Aware 3D-LLM}
\label{sec:cairn}
\begin{figure}
    \centering
    \includegraphics[width=\linewidth]{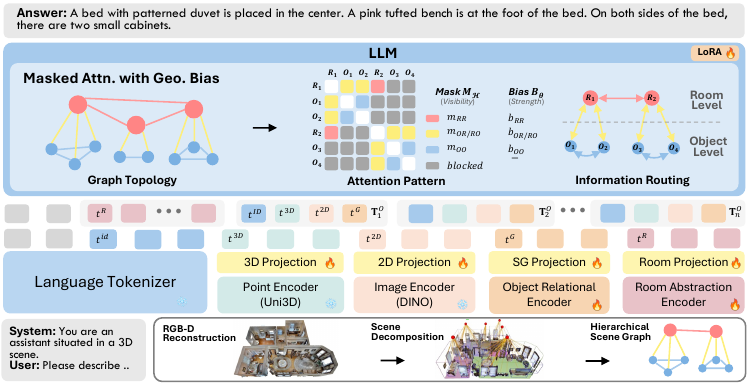}
    \vspace{-6mm}
    \caption{\textbf{Overview of CAIRN.} Given a 3D scene, CAIRN constructs a hierarchical scene graph capturing object relations and room adjacencies (bottom), tokenizes it into object and room tokens (middle), and feeds them to an LLM with hierarchical masked attention and geometric bias (top). The mask routes information along scene topology, while learned bias terms inject spatial priors into attention logits, enabling topology-aware reasoning across rooms.}
    \label{fig:framework}
    \vspace{-1.2em}
\end{figure}

Our goal is to enable 3D-LLMs to perform multi-room 3D scene understanding in full household environments. Our approach, \textbf{CAIRN}, explicitly models the hierarchical topology of multi-room 3D scenes, enabling 3D-LLMs to reason over both object-level and room-level structure, as illustrated in~\Cref{fig:framework}. We first describe how we represent a multi-room 3D scene as a hierarchical 3D scene graph, capturing both object relations and room adjacencies (\Cref{sec:representation}). We then describe how this graph is encoded into scene tokens while preserving both object-level and room-level information (\Cref{sec:tokenization}). Finally, we detail how CAIRN incorporates structured masked attention and geometric bias to make 3D-LLMs topology-aware by constraining information flow according to scene connectivity and injecting explicit spatial priors into attention (\Cref{sec:attention}).

\subsection{Multimodal Hierarchical Scene Graph Construction}
\label{sec:representation}
CAIRN represents a multi-room 3D scene as a two-layer hierarchical scene graph $\mathcal{G}_{\mathcal{H}}$. The lower layer encodes object-level structure $(\mathcal{V}_O,\mathcal{E}_{OO})$, while the upper layer captures room-level topology $(\mathcal{V}_R,\mathcal{E}_{RR})$. Cross-layer edges $\mathcal{E}_{OR}$ assign each object to its containing room, forming a graph as:
\begin{equation}
\mathcal{G}_{\mathcal{H}}
=
\big(
(\mathcal{V}_O,\mathcal{E}_{OO}),
\mathcal{E}_{OR},
(\mathcal{V}_R,\mathcal{E}_{RR})
\big).
\label{eq:hsg}
\end{equation}
The graph construction is agnostic to the scene source, and can be derived from RGB-D reconstructions or simulator annotations. For the object-level graph, we follow Chat-Scene~\cite{chatscene} and obtain object instances using Mask3D~\cite{mask3d}. Each instance defines a node $\mathcal{O}_i \in \mathcal{V}_O$ with associated geometry and visual features. Edges $\mathcal{E}_{OO}$ connect each object to its top-$K_g$ nearest neighbors within the same room, with attributes encoding relative geometry and semantic relations~\cite{wang2023vl}.

For the room-level graph, room nodes $\mathcal{V}_R=\{\mathcal{R}_r\}_{r=1}^{M}$ are obtained via bird's-eye-view occupancy partitioning~\cite{werby2024hierarchical}, and adjacent rooms are connected by edges $\mathcal{E}_{RR}$. Object-room edges $\mathcal{E}_{OR}$ are induced by assigning each object to its corresponding room. The resulting hierarchical graph captures both local object arrangement and global scene topology, and is tokenized for LLM reasoning.

\subsection{Hierarchical Scene Graph Tokenization}
\label{sec:tokenization}
\textbf{Object-Level Tokenization.}
To bridge multimodal object information to the LLM token space, each object node $\mathcal{O}_i\in\mathcal{V}_O$ is mapped to a fixed token block
\begin{equation}
\mathbf{T}_i^O =
[t_i^{\mathrm{id}},\, t_i^{3D},\, t_i^{2D},\, t_i^{G}],
\qquad
t_i^{\mathrm{id}}, t_i^{3D}, t_i^{2D}, t_i^{G} \in \mathbb{R}^{d},
\label{eq:obj_tok}
\end{equation}
which encodes identity, 3D geometry, 2D appearance, and relational context. The identifier token $t_i^{\mathrm{id}}$ is obtained from a unique textual tag (e.g. \texttt{<OBJ\_01>}) via the LLM tokenizer, while visual tokens are computed by projecting pretrained 3D and 2D features into the LLM space.

The graph token $t_i^{G}$ encodes relational context beyond per-object geometry and appearance, capturing spatial interactions between objects. To this end, we apply $L_g$ layers of message passing over object-object edges $\mathcal{E}_{OO}$, where each object updates its representation by aggregating information from connected neighbors. The resulting representation is projected into the LLM space to obtain $t_i^{G}$. See Appendix~\ref{app:appendix_arch} for more details.

\textbf{Room-Level Tokenization.}
While room nodes define the higher-level semantic structure of the scene, geometry alone is insufficient to capture semantic room content. We therefore obtain room tokens by aggregating object-level tokens within each room. For each room $\mathcal{R}_r$, we introduce $K_r$ learned queries $\{q_{r,k}\}_{k=1}^{K_r}$ that cross-attend to object tokens within the room:
\begin{equation}
t_{r,k}^{R} = W_R \,\mathrm{CrossAttn}
\left(
q_{r,k},
\{t_i\}_{i\in\mathcal{O}^{(r)}}
\right).
\label{eq:room_tok}
\end{equation}
Resulting room tokens $\mathbf{T}_r^R=[t_{r,1}^{R},\ldots,t_{r,K_r}^{R}]$ summarize room-level content and enable cross-room communication. Object-room assignments and room adjacency are used for topology-aware attention.

\subsection{Hierarchical Masked Attention with Geometric Bias}
\label{sec:attention}

In our hierarchical scene graph $\mathcal{G}_{\mathcal{H}}$ (\Cref{sec:representation}), the scene exhibits both object- and room-level structure. Standard self-attention ignores this structure, treating all tokens as fully connected. We instead formulate attention as information routing over the graph: a structured mask determines which routes are permitted, while a geometric bias modulates the strength of information flow along them. This design encodes scene topology and spatial priors directly into the attention mechanism, rather than learning them implicitly from data.

Let $\mathbf{T}=[\mathbf{T}_{\mathrm{lang}};\mathbf{T}_{\mathrm{scene}}]\in\mathbb{R}^{L\times d}$ denote the LLM input, where $\mathbf{T}_{\mathrm{lang}}$ contains prompt tokens and $\mathbf{T}_{\mathrm{scene}}$ contains object and room tokens. The CAIRN attention operator is defined as
\begin{equation}
\mathrm{Attn}(X)
=
\mathrm{Softmax}\!\left(
\frac{QK^\top}{\sqrt{d}}
+ M_{\mathcal{H}}
+ B_{\theta}
\right)V.
\label{eq:attn}
\end{equation}
The two additive terms play complementary roles: $M_{\mathcal{H}}$ enforces which token pairs may interact, while $B_{\theta}$ modulates their interaction strength based on geometry. The mask is applied only among object and room tokens (prompt tokens retain standard attention), with allowed pairs defined as
\begin{equation}
\mathcal{S}_{\mathcal{H}}
=
\mathcal{S}_{\mathrm{OO}}
\cup
\mathcal{S}_{\mathrm{OR}}
\cup
\mathcal{S}_{\mathrm{RR}}.
\label{eq:mask_set}
\end{equation}
Each subset corresponds to edges in $\mathcal{G}_{\mathcal{H}}$ (\Cref{eq:hsg}) and defines bidirectional token interactions. This induces a hierarchical routing pattern: object tokens interact only within their room, and cross-room communication is mediated through room tokens, which carry inter-room information flow.

To modulate the strength of admitted interactions, the geometric bias adds a learned offset to the attention logits:
\begin{equation}
    B_{\theta}(u,v)=b_{\tau(u,v)}(\mathbf{g}_{uv}),
    \label{eq:bias}
\end{equation}
where $\tau(u,v)\in\{\mathrm{OO},\mathrm{OR},\mathrm{RR}\}$ denotes the interaction type and $b_{\tau}$ is a type-specific MLP. The descriptor $\mathbf{g}_{uv}$ captures the spatial relation at each level: relative position and distance for OO pairs, object-to-room location for OR pairs, and relative layout for RR pairs.

$B_{\theta}$ is applied to the first $L_b$ layers and later layers only retain the structural sparsity induced by $M_{\mathcal{H}}$. This yields a block-sparse attention pattern (\Cref{fig:framework}) that reduces object-object attention from $O(N^2)$ to $O\!\left(\sum_r |\mathcal{O}^{(r)}|^2\right)$. Together, $M_{\mathcal{H}}$ and $B_{\theta}$ turn attention into a structural operator over the scene hierarchy, where topology determines token visibility and geometry determines interaction strength.

\section{Experiments}\label{sec:exp}
We evaluate CAIRN through four questions: 1) How is the quality of our constructed CAIRN-MR benchmark? 2) Does the hierarchical design of CAIRN 3D-LLM improve performance on cross-room reasoning tasks? 3) Does CAIRN remain effective on single-room scene understanding? 4) Which components of CAIRN are responsible for the observed gains, and for which task types?

\subsection{Implementation Details}
We use Qwen3-8B~\cite{qwen3} as the LLM backbone, frozen and adapted via LoRA~\cite{lora} (rank 16, all attention and MLP projections); input embedding and output head are unfrozen for object identifier tokens. The graph encoder uses $L_g{=}2$ message-passing layers, each room is summarized by $K_r{=}4$ learned queries, and geometric bias $B_{\theta}$ is applied to the first $L_b{=}4$ transformer layers. Training has two stages on $8{\times}$H200 GPUs (${\sim}$12 hours total): Stage~1 trains for 3 epochs on ScanNet tasks~\cite{scanrefer, multi3drefer, scan2cap, scanqa, sqa3d}; Stage~2 trains for 3 epochs on 30\% ScanNet and 100\% CAIRN-MR. Full hyperparameters and training details are provided in Appendix~\ref{app:appendix_hyperparams}.

\subsection{Experiment Setup}

\textbf{Datasets.}
For single-room evaluation, we use five ScanNet-based benchmarks: ScanRefer~\cite{scanrefer} and Multi3DRefer~\cite{multi3drefer} (grounding), Scan2Cap~\cite{scan2cap} (captioning), ScanQA~\cite{scanqa} (VQA), and SQA3D~\cite{sqa3d} (situated QA). For multi-room evaluation, we use CAIRN-MR (\Cref{sec:tasks}). Splits are partitioned at the building level with no scene overlap. 

\textbf{Metrics.}
We follow standard protocols for ScanNet benchmarks: Acc@0.25/0.5 for ScanRefer, F1@0.25/0.5 for Multi3DRefer, CIDEr@0.5 and BLEU-4@0.5 for Scan2Cap, CIDEr and BLEU-4 for ScanQA, and exact match for SQA3D. For CAIRN-MR, grounding and captioning follow ScanRefer and Scan2Cap protocols. QA tasks use rule-based answer extraction, with malformed outputs scored as incorrect: Identification and Comparison are evaluated by option selection, Existence by binary accuracy, and Counting by Mean Relative Accuracy (MRA)~\cite{vsibench}.


\subsection{Benchmark Quality}
We evaluate the quality of CAIRN-MR's QA tasks along three dimensions: annotation reliability, task difficulty, and answer distribution, using three baselines (\Cref{tab:calibration}): Random (selects answers uniformly), Frequency (predicts the most frequent answer), and Human (3 annotators on a subset). We focus on QA tasks as they provide discrete, verifiable outputs for controlled calibration.

\textbf{Annotation Quality.} Human performance is consistently high across tasks (86.8\%--93.5\%), indicating that the annotations are reliable and that the tasks are well-defined. This suggests that performance gaps primarily reflect task difficulty rather than annotation noise.

\begin{wraptable}{r}{0.49\textwidth}
\vspace{-10pt}
\centering
\small
\setlength{\tabcolsep}{4pt}
\caption{\textbf{Calibration of QA tasks in CAIRN-MR.} Performance of random, frequency, and human baselines across tasks.}
\label{tab:calibration}
\begin{tabular}{lccccc}
\toprule
             & Ident. & \multicolumn{2}{c}{Count.} & Exist. & Comp. \\
\cmidrule(lr){3-4}
             & EM$\uparrow$ & EM$\uparrow$ & MRA$\uparrow$ & EM$\uparrow$ & EM$\uparrow$ \\
\midrule
Random       & 25.0 & 11.1 & 19.9 & 50.0 & 50.0 \\
Frequency    & 25.3 & 54.7 & 62.6 & 50.0 & 48.9 \\
Human        & 86.8 & 82.3 & 94.5 & 93.5 & 91.4 \\
\bottomrule
\end{tabular}
\vspace{-10pt}
\end{wraptable}

\textbf{Task Difficulty.}
Human performance substantially exceeds random and frequency baselines across all tasks, indicating that the tasks are non-trivial and require genuine spatial reasoning beyond surface statistics.

\textbf{Answer Distribution Balance.}
Random and frequency baselines are closely aligned on Identification, Existence, and Comparison, suggesting balanced answer distributions without dominant labels. Counting differs due to skew toward small integers (e.g., 2 accounts for 49.8\%), allowing strong frequency performance (62.6\% MRA) from numerical priors rather than reasoning. This skew is intrinsic to indoor scenes~\cite{scanqa} and serves as a reference for interpreting the metric.

\subsection{Performance on Multi-Room Scene Understanding}
\Cref{tab:hm3d_main} compares CAIRN with three representative baselines on CAIRN-MR, all trained under the same two-stage protocol with Qwen3-8B.
CAIRN achieves the largest gains on grounding and captioning: +5.4 Acc@0.25/+5.4 Acc@0.5 and +14.9 CIDEr (roughly 19\% relative). These improvements reflect more accurate cross-room disambiguation enabled by topology-aware modeling of room-level structure.

Across QA tasks, gains scale with reliance on room-level structure. Tasks requiring cross-room reasoning (Comparison, Existence) show the largest improvements (+6.7, +3.4 EM), as they depend on reasoning across room boundaries. Counting, which enumerates objects within a localized room, shows moderate gains (+2.2 MRA), benefiting from clearer room-level scope. Identification, which only requires recognizing an object within an already localized room, remains comparable across methods (62.1 vs.\ 61.9 EM). This trend, from cross-room reasoning to within-room enumeration to object recognition, shows that CAIRN’s gains scale with the level of room-level abstraction required. The improvements are consistent across backbones with strong performance even for smaller models, showing that CAIRN remains effective under reduced model capacity. (\Cref{tab:hm3d_main}).
\begin{table}[t]
\caption{\textbf{CAIRN-MR multi-room performance comparison.} Results on grounding, captioning, and question answering tasks. Best results are in \textbf{bold}.}
\centering
\small
\resizebox{\linewidth}{!}{
\begin{tabular}{l c cc cc cccc}
\toprule
\multirow{2}{*}{Method} &
\multirow{2}{*}{LLM} &
\multicolumn{2}{c}{Grounding} &
\multicolumn{2}{c}{Captioning} &
\multicolumn{4}{c}{Question Answering} \\
\cmidrule(lr){3-4}
\cmidrule(lr){5-6}
\cmidrule(lr){7-10}
& &
\makecell{A@\\0.25$\uparrow$} & \makecell{A@\\0.5$\uparrow$} &
C$\uparrow$ & B-4$\uparrow$ &
\makecell{Ident.\\EM$\uparrow$} &
\makecell{Count.\\MRA$\uparrow$} &
\makecell{Exist.\\EM$\uparrow$} &
\makecell{Comp.\\EM$\uparrow$} \\
\midrule
Chat-Scene~\cite{chatscene} &
Qwen3-8B &
19.6 & 19.4 &
78.3 & 21.2 &
61.3 & 58.9 & 69.5 & 63.6 \\
3DGraphLLM~\cite{3dgraphllm} &
Qwen3-8B &
18.9 & 18.7 &
79.6 & 21.1 &
61.9 & 58.3 & 72.7 & 62.3 \\
Inst3D-LMM~\cite{yu2025inst3d} &
Qwen3-8B &
20.4 & 20.0 &
79.5 & 21.3 &
61.8 & 58.5 & 71.6 & 63.7 \\
\midrule
CAIRN (ours) &
Llama-3.2-3B &
24.2 & 23.8 &
91.0 & 22.2 &
60.9 & 59.8 & 75.5 & 70.0 \\
 &
Llama-3.1-8B &
24.6 & 24.1 &
89.5 & 21.9 &
\textbf{62.5} & 59.4 & 75.9 & 69.8 \\
  &
Qwen3-4B &
24.3 & 23.9 &
89.7 & 22.0 &
61.9 & 60.8 & 75.7 & 68.2 \\
\rowcolor{gray!8}
  &
Qwen3-8B &
\textbf{25.8} & \textbf{25.4} &
\textbf{94.5} & \textbf{22.8} &
62.1 & \textbf{61.1} & \textbf{76.1} & \textbf{70.4} \\
\bottomrule
\end{tabular}
}
\label{tab:hm3d_main}
\end{table}
\begin{wrapfigure}{r}{0.5\linewidth}
\vspace{-1.0em}
  \centering
  \includegraphics[width=1\linewidth]{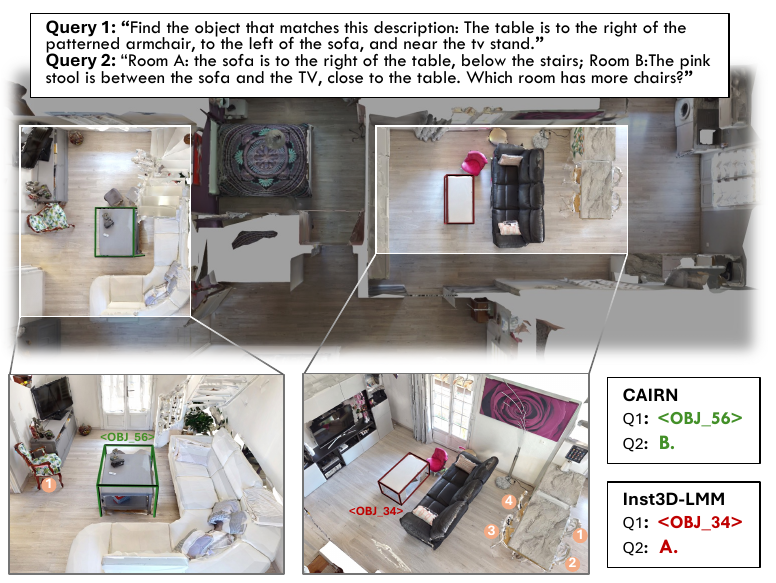}
  \vspace{-5mm}
  \caption{\textbf{Qualitative results on a multi-room scene} with visually similar living rooms. Green denotes CAIRN; red denotes the best baseline.}
  \label{fig:wrap}
  \vspace{-5.0em}
\end{wrapfigure}

\subsection{Performance on Single-Room Setting}

\begin{table}[t]
\caption{\textbf{Single-room performance comparison.} Results on grounding, captioning, and question answering tasks. Best in \textbf{bold}; second best \underline{underlined}.}
\centering
\small
\setlength{\tabcolsep}{4pt}
\renewcommand{\arraystretch}{1.12}
\resizebox{\textwidth}{!}{
\begin{tabular}{p{0.25cm}l c cc cc cc cc c}
\toprule
& \multirow{2}{*}{Method} & \multirow{2}{*}{LLM} &
\multicolumn{2}{c}{ScanRefer} &
\multicolumn{2}{c}{Multi3DRefer} &
\multicolumn{2}{c}{Scan2Cap} &
\multicolumn{2}{c}{ScanQA} &
\multicolumn{1}{c}{SQA3D} \\
\cmidrule(lr){4-5}
\cmidrule(lr){6-7}
\cmidrule(lr){8-9}
\cmidrule(lr){10-11}
\cmidrule(lr){12-12}
& & &
A@0.25$\uparrow$ & A@0.5$\uparrow$ &
F1@0.25$\uparrow$ & F1@0.5$\uparrow$ &
C@0.5$\uparrow$ & B-4@0.5$\uparrow$ &
C$\uparrow$ & B-4$\uparrow$ &
EM$\uparrow$ \\
\midrule
\multirow{7}{*}{\centering\rotatebox{90}{Expert Models}}
& 3DVG-Trans~\cite{3dvg-transformer}       & $\times$ & 45.9 & 34.5 & -    & -    & -    & -    & -    & -    & -    \\
& ViL3DRel~\cite{vil3drel}                 & $\times$ & 47.9 & 37.7 & -    & -    & -    & -    & -    & -    & -    \\
& M3DRef-CLIP~\cite{multi3drefer}          & $\times$ & 51.9 & 44.7 & 42.8 & 38.4 & -    & -    & -    & -    & -    \\
& 3D-VLP~\cite{3d-vlp}                     & $\times$ & 51.4 & 39.5 & -    & -    & 54.9 & 32.3 & 67.0 & 11.1 & -    \\
& 3D-VisTA~\cite{3dvista}                  & $\times$ & 50.6 & 45.8 & -    & -    & 66.9 & 34.0 & 72.9 & 13.1 & 48.5 \\
& Vote2Cap-DETR++~\cite{vote2cap++}        & $\times$ & -    & -    & -    & -    & 67.6 & 37.1 & -    & -    & -    \\
& PQ3D~\cite{pq3d}                         & $\times$ & -    & 51.2 & 50.1 & -    & 80.3 & 36.0 & 87.8 & -    & 47.1 \\
\midrule
\multirow{12}{*}{\centering\rotatebox{90}{LLM-based Models}}
& 3D-LLM~\cite{3dllm}                      & Flamingo        & 21.2 & -    & -    & -    & -    & -              & 59.2          & 7.2           & -    \\
& Scene-LLM~\cite{scenellm}                & Llama-2-7B      & -    & -    & -    & -    & -    & -              & 80.0          & 12.0          & 54.2 \\
& LL3DA~\cite{ll3da}                       & OPT-1.3B        & -    & -    & -    & -    & 65.2 & 36.8           & 76.8          & 13.5          & -    \\
& LEO~\cite{leo}                           & Vicuna-7B-v1.1  & -    & -    & -    & -    & 72.4 & \underline{38.2} & \textbf{101.4} & 13.2         & 50.0 \\
& Grounded 3D-LLM~\cite{grounded3dllm}     & Tiny-Vicuna-1B  & 47.9 & 44.1 & 45.2 & 40.6 & 70.6 & 35.5           & 72.7          & 13.4          & -    \\
& GPT4Scene-HD~\cite{qi2025gpt4scene}      & Qwen2-VL-7B     & 50.9 & 46.4 & 53.7 & 50.0 & 74.4 & 37.9           & 89.9          & \textbf{15.9} & \underline{57.2} \\
& Chat-Scene~\cite{chatscene}              & Vicuna-7B-v1.5  & 55.5 & 50.2 & 57.1 & 52.4 & 77.1 & 36.3           & 87.7          & 14.3          & 54.6 \\
&                                          & Qwen3-8B        & 57.5 & 51.9 & 61.3 & 56.7 & 75.4 & 34.5           & 90.3          & 13.1          & 55.8 \\
& 3DGraphLLM~\cite{3dgraphllm}             & Qwen3-8B        & 58.4 & \underline{53.1} & \underline{62.9} & \underline{58.1} & 78.1 & 35.2 & 91.2 & 13.1 & 56.1 \\
& Inst3D-LMM~\cite{yu2025inst3d}              & Vicuna-7B-v1.5  & 57.8 & 51.6 & 58.3 & 53.5 & 79.7 & \textbf{38.3}  & 88.6          & \underline{14.9} & -    \\
&                                          & Qwen3-8B        & \underline{58.9} & 52.7 & 62.4 & 57.9 & \textbf{80.9} & 38.0 & 91.5 & 13.6 & -    \\
\rowcolor{gray!8}
& CAIRN (ours)                             & Qwen3-8B        & \textbf{59.6} & \textbf{54.1} & \textbf{63.8} & \textbf{58.8} & \underline{80.7} & 36.2 & \underline{92.1} & 13.5 & \textbf{58.1} \\
\bottomrule
\end{tabular}
}
\label{tab:benchmark}
\end{table}
\Cref{tab:benchmark} reports results on five ScanNet-based single-room benchmarks. 
Since these scenes contain a single room, the hierarchical structure reduces to a single level, leaving graph tokens and geometric bias as the effective components. CAIRN matches or exceeds prior methods across all benchmarks, achieving the best or second-best results on 7 of 9 metrics. These results indicate that CAIRN’s room-level components do not interfere with single-room performance. The gains are primarily driven by graph tokens and geometric bias, which encode local relational structure independent of room hierarchy, yielding improved grounding performance while maintaining competitive results across tasks.

\subsection{Ablation Studies}
\label{sec:ablation}
\Cref{tab:hm3d_ablation} presents an ablation on CAIRN-MR to analyze component contributions. Starting from the full model, we remove topology-aware attention (hierarchical mask and geometric bias) and structured representations (room and graph tokens), resulting in a flat baseline.
\begin{table}[t]
\caption{\textbf{Ablation of CAIRN.} We start from the full model and progressively remove components: geometric attention bias, hierarchical attention mask, room-level tokens, and graph tokens.}
\centering
\small
\setlength{\tabcolsep}{3.5pt}
\renewcommand{\arraystretch}{1.1}
\resizebox{\linewidth}{!}{
\begin{tabular}{l c c c c cc cc cccc}
\toprule
\multirow{2}{*}{Variant} &
\multirow{2}{*}{GNN} &
\multirow{2}{*}{Room} &
\multirow{2}{*}{Mask} &
\multirow{2}{*}{Bias} &
\multicolumn{2}{c}{Grounding} &
\multicolumn{2}{c}{Captioning} &
\multicolumn{4}{c}{Question Answering} \\
\cmidrule(lr){6-7}
\cmidrule(lr){8-9}
\cmidrule(lr){10-13}
& & & & &
\makecell{A@\\0.25$\uparrow$} & \makecell{A@\\0.5$\uparrow$} &
C$\uparrow$ & B-4$\uparrow$ &
\makecell{Ident.\\EM$\uparrow$} &
\makecell{Count.\\MRA$\uparrow$} &
\makecell{Exist.\\EM$\uparrow$} &
\makecell{Comp.\\EM$\uparrow$} \\
\midrule
\rowcolor{gray!8}
CAIRN (full)            & \checkmark & \checkmark & \checkmark & \checkmark & \textbf{25.8} & \textbf{25.4} &
\textbf{94.5} & \textbf{22.8} &
\textbf{62.1} & \textbf{61.1} & \textbf{76.1} & \textbf{70.4} \\
\midrule
$-$ geometric bias      & \checkmark & \checkmark & \checkmark & --         & 24.5 & 24.1 & 92.4 & 22.4 & 62.1 & 60.6 & 74.5 & 69.8 \\
$-$ hierarchical mask   & \checkmark & \checkmark & --         & --         & 21.6 & 21.1 & 85.7 & 21.8& 61.8 & 59.3 & 72.6 & 64.3 \\
$-$ room tokens         & \checkmark & --         & --         & --         & 20.8 & 20.6 & 82.3 & 21.5 & 62.3 & 57.9 & 73.0 & 62.2 \\
$-$ graph tokens (= flat)  & --         & --         & --         & --         & 19.6 & 19.4 & 78.3 & 21.2 & 61.3 & 58.6 & 69.5 & 63.6 \\
\bottomrule
\end{tabular}
}
\label{tab:hm3d_ablation}
\end{table}

\noindent \textbf{Effectiveness of Topology-Aware Attention.}
Topology-aware attention is the primary driver of performance gains. Removing the hierarchical mask causes the largest drops across all metrics (e.g., $-5.5$ EM on comparison, $-6.7$ CIDEr on captioning, $-2.9$ Acc@0.25 on grounding), as it enforces topology-aware routing of cross-room information. Without it, cross-room information flows become unconstrained, diluting room-level distinctions even when room tokens are present. The geometric bias provides consistent but smaller gains (e.g., $+1.3$ Acc@0.25, $+1.6$ EM) by injecting structure into attention.

\noindent \textbf{Effectiveness of Structured Representations.}
Structured representations provide complementary gains by encoding global and local context. Room tokens aggregate room-level information into compact summaries, improving captioning ($+3.4$ CIDEr) and comparison ($+2.1$ EM), while graph tokens encode local object relations, contributing gains across captioning ($+4.0$ CIDEr), existence ($+3.5$ EM), and grounding ($+1.2$ Acc@0.25), particularly in tasks where local context supports spatial reasoning.

\noindent \textbf{Gains Increase with Room-Level Reasoning.} Performance gains increase with the degree of room-level reasoning required. Identification remains largely unaffected (within 1 point), as it primarily relies on recognizing objects within a localized room. Counting shows moderate degradation ($\sim$3 points), reflecting its dependence on aggregating instances within a correctly identified room. In contrast, Existence and Comparison exhibit the largest drops ($\sim$7--8 points), as they require reasoning across rooms and are more sensitive to disruptions in cross-room information flow. This pattern mirrors the graded trend observed in~\Cref{tab:hm3d_main}.

\section{Conclusion}
\label{sec:conclusion}
We introduce CAIRN, a topology-aware 3D-LLM for multi-room scene understanding, together with CAIRN-MR, a benchmark requiring both object- and room-level reasoning. By modeling hierarchical scene structure, CAIRN aligns information flow with scene topology via masked attention and encodes spatial priors through geometric bias. Experiments show that CAIRN consistently outperforms prior methods across grounding, captioning, and question answering on CAIRN-MR, while generalizing effectively to single-room benchmarks. This suggests that topology-aware modeling enables structured 3D reasoning across complex multi-room environments.

\textbf{Limitations and Future Work.} Further scaling model capacity and training data may yield additional gains. Extending CAIRN-MR to more complex, multi-level environments would broaden the scope of structured reasoning. Another direction is to incorporate explicit reasoning processes for more interpretable multi-step reasoning over scene structure, potentially improving compositional generalization across more complex scene configurations.
\newpage

{
    \small
    \bibliographystyle{plainnat}
    \bibliography{references}
}
\clearpage
\appendix
\begin{center}
    {\Large\bfseries Appendix for CAIRN\par}
\end{center}
\addcontentsline{toc}{section}{Appendix for CAIRN}
\vspace{1em}
\section{Overview}
This appendix provides supplementary materials supporting the main paper. \Cref{app:bench_construction} details the construction of the CAIRN-MR benchmark, including multi-room scene assembly from HM3D~\cite{ramakrishnan2021habitat}, the procedure for generating and validating spatial referring expressions, appearance-based noun augmentation via vision-language models, and the instantiation of grounding, captioning, and four QA task formats. \Cref{app:appendix_implementation} provides additional implementation details, including model hyperparameters and architectural details for object-level tokenization.

\section{Benchmark Construction Details}
\label{app:bench_construction}
\subsection{Multi-Room Scene Assembly}
\label{app:scene_assembly}
CAIRN-MR is built on HM3D~\cite{ramakrishnan2021habitat}, a large-scale collection of realistic residential environments. 
We adopt the per-room annotations provided by SceneVerse~\cite{jia2024sceneverse}, which include object identities, semantic labels, instance segmentations, and pairwise spatial relations for each room in the HM3D scenes.
To construct multi-room scenes, we use the ground-truth room positions provided in HM3D.
For each scene, we select a seed room whose spatial extent exceeds 30\,m$^2$, then compute Euclidean distances from its center to all other rooms and select the 3--5 nearest rooms to form a multi-room layout.
To ensure spatial connectivity, we obtain the ground-truth occupancy map between sampled rooms via Habitat simulator~\cite{puig2023habitat} and discard any sample that contains an isolated room unreachable from the rest.
Since different seed rooms within the same base scene yield different multi-room combinations, a single HM3D scene can produce multiple distinct samples.
The resulting benchmark comprises 479 training samples from 116 base scenes and 194 validation samples from 45 base scenes, partitioned at the building level with no scene overlap.

\subsection{Referring Expression Construction}
\label{app:ref_construction}
We construct object-level referring expressions from the pairwise spatial relations provided by SceneVerse~\cite{jia2024sceneverse}, which cover 11 relation types including support, containment, proximity, and directional positioning (see~\Cref{tab:relations} for the full inventory).
Each relation is a triplet $(o_i, r, o_j)$ linking two objects within the same room.
From these triplets, we construct referring expressions in three forms of increasing structural complexity.

\begin{wraptable}{r}{0.49\textwidth}
\centering
\small
\caption{Spatial relation types used for referring expression construction.}
\label{tab:relations}
\begin{tabular}{ll}
\toprule
Relation type & Example phrases \\
\midrule
Support        & resting on, placed on \\
Embedded       & embedded into \\
Inside         & inside \\
Hanging        & hanging on, mounted on \\
Above / Below  & above, below \\
Directional (near)  & to the left of, behind \\
Directional (far)   & far from \\
Proximity      & close to, next to \\
Aligned        & aligned with \\
In the middle of & in the middle of \\
\bottomrule
\end{tabular}
\end{wraptable}

\noindent\textbf{1) Multi-reference.}
For a target object, we retrieve 2--3 relations that each directly involve the target, and combine them into a single expression via conjunction templates.
SceneVerse's multi-object relations (e.g., aligned, in the middle of) naturally fall into this category.
Example: ``the table \textbf{between} the sofa \textbf{and} the bookshelf, \textbf{next to} the lamp.''

\noindent\textbf{2) Multi-hop chain.}
We identify a chain of three consecutive relations within the same room, where the target object appears at any position along the chain.
The chain provides progressively more spatial context for disambiguation.
Example: ``the chair \textbf{next to} the desk \textbf{that supports} the monitor \textbf{near} the window.''

\noindent\textbf{3) Chain with branch.}
We first identify a relation chain of two or more hops (as in form 2), then connect the target object to any node on the chain via an additional relation.
The chain serves as room-level context, while the branch relation localizes the target.
Example: ``the lamp \textbf{above} the nightstand, where the nightstand \textbf{is next to} the bed \textbf{that faces} the window.''

\vspace{4pt}

Prior to expression construction, we exclude objects with structurally ambiguous or overly generic labels (e.g., wall, floor, ceiling, object, furniture) and objects with fewer than 128 surface points in the reconstructed point cloud, ensuring that all participating instances are semantically specific and geometrically well-reconstructed.
After construction, each expression is verified against the full multi-room scene to ensure it uniquely identifies the intended target.
Specifically, for each expression targeting an object of class $c$ with spatial constraints $\mathcal{C} = \{(r_k, c_k)\}_{k=1}^{K}$ (where each constraint specifies a relation type $r_k$ and a reference object class $c_k$), we enumerate all objects of class $c$ across the entire scene and check whether each candidate simultaneously satisfies all constraints in $\mathcal{C}$.
An expression is retained only if exactly one object---the intended target---matches; all others are discarded.
Since this verification is performed on the assembled multi-room scene, uniqueness is guaranteed at the full-scene level rather than within a single room. The verification procedure is formally described in Algorithm~\ref{alg:expr_verify}.

\begin{algorithm}[H]
\SetAlgoLined
\DontPrintSemicolon
\SetKwInOut{Input}{Input}
\SetKwInOut{Output}{Output}
\SetKwFunction{Satisfies}{Satisfies}
\SetKwFunction{Class}{Class}

\Input{Multi-room scene $\mathcal{S}$ with object set $\mathcal{O}$;\\
       candidate expression $e$ targeting object $o^\star \in \mathcal{O}$ of class $c$;\\
       spatial constraints $\mathcal{C} = \{(r_k, c_k)\}_{k=1}^{K}$ parsed from $e$\;}
\Output{Boolean $\textsc{Retain}(e) \in \{\text{true}, \text{false}\}$\;}
\BlankLine

$\mathcal{M} \leftarrow \emptyset$ \tcp*{matches across the full scene}
\BlankLine

\ForEach{$o \in \mathcal{O}$ \textbf{with} $\Class(o) = c$}{
    $\textit{ok} \leftarrow \text{true}$\;
    \ForEach{$(r_k, c_k) \in \mathcal{C}$}{
        \If{$\nexists\, o' \in \mathcal{O}$ \textbf{with} $\Class(o') = c_k$ \textbf{and} $\Satisfies(o, r_k, o')$}{
            $\textit{ok} \leftarrow \text{false}$\;
            \textbf{break}\;
        }
    }
    \lIf{\textit{ok}}{$\mathcal{M} \leftarrow \mathcal{M} \cup \{o\}$}
}
\BlankLine

\eIf{$|\mathcal{M}| = 1$ \textbf{and} $\mathcal{M} = \{o^\star\}$}{
    \Return \textnormal{true} \tcp*{unique match: retain expression}
}{
    \Return \textnormal{false} \tcp*{ambiguous or wrong match: discard}
}
\caption{Full-scene uniqueness verification for a referring expression.}
\label{alg:expr_verify}
\end{algorithm}

\subsection{Appearance-Based Noun Augmentation}
\label{app:appearance}

To increase linguistic diversity, we optionally augment object nouns with visually grounded appearance modifiers.
This augmentation operates on a per-object basis and proceeds as follows.

For each object instance, we render the assembled multi-room scene and select multi-view RGB frames in which the object is visible, based on its 3D position and camera coverage.
We then run SAM~3~\cite{carion2025sam3segmentconcepts} on each frame using the object's category label as the text prompt, and retain only frames in which the object is detected with confidence above 0.7.
Objects with fewer than 3 qualifying frames are excluded from augmentation entirely, as insufficient high-confidence views indicate either poor reconstruction quality or heavy occlusion.
For objects with more than 5 qualifying frames, we retain the 5 frames with the highest detection confidence to limit computational cost.
The detected bounding boxes are used to crop tightly around the object in each selected frame, producing a set of 3--5 cropped views per eligible object.

For each eligible object, we independently query 
Qwen3-VL-32B~\cite{qwen3} on each of its 3--5 
cropped views. The model is provided with the object's 
category label and asked to identify a single clearly 
visible attribute from three types: color, material, 
or distinctive part. The model may also output 
``none'' if no attribute is clearly identifiable. 
The full prompt template is shown in~\Cref{fig:vlm_prompt}.

\begin{figure}[h]
\centering
\begin{tcolorbox}[
  colback=gray!3,
  colframe=black!40,
  boxrule=0.4pt,
  width=0.95\linewidth,
  arc=2pt,
  left=8pt,
  right=8pt,
  top=6pt,
  bottom=6pt,
  title={\small Prompt Template for Appearance Augmentation},
  fonttitle=\bfseries\small
]
\small\ttfamily
You are given a cropped image of a \{category\}.
Select ONE visual attribute that is clearly visible:\\[4pt]
- color (e.g., white, brown, black)\\
- material (e.g., wooden, metal, glass)\\
- shape (e.g., round, rectangular, L-shaped)\\[4pt]
If none is clearly identifiable, output: none\\
Output format: [attribute] \{category\}\\
Examples: wooden chair, black table, round rug, none
\end{tcolorbox}
\caption{Prompt template for querying Qwen3-VL-32B. \texttt{\{category\}} is replaced with the object's semantic label at runtime.}
\label{fig:vlm_prompt}
\end{figure}

To suppress hallucination, we apply multi-view 
consensus: an attribute is accepted only if it appears 
in at least half of the queried views for that object. 
Objects for which no attribute achieves consensus 
retain their original noun without modification. 
Under this procedure, 21.6\% of object nouns in the 
benchmark are augmented with an appearance modifier.
The augmentation does not alter target identity or 
spatial constraints in the associated expressions.

\subsection{Task Instantiation}
\label{app:task_instantiation}

All benchmark tasks are instantiated from the validated referring expression pool described in~\Cref{app:ref_construction}. 
We partition the expression pool into two equal halves per scene using a fixed random seed: one half is used for grounding, the other for captioning.

\paragraph{Grounding.}
Each sample maps a referring expression to its target object identifier.
Queries are drawn from three prompt templates, where \texttt{\{description\}} is replaced by the referring expression at runtime:
\begin{quote}\small
\texttt{``Find the object that matches this description: "\{description\}"''} \\
\texttt{``Which object in the scene is being described? "\{description\}"''} \\
\texttt{``Identify the object being referred to: "\{description\}"''}
\end{quote}
The expected answer is the object identifier (e.g., \texttt{<OBJ041>}).

\paragraph{Captioning.}
Each sample maps an object identifier to a referring expression.
Queries are drawn from four prompt templates, where \texttt{\{id\}} is replaced by the object identifier:
\begin{quote}\small
\texttt{``Explain the spatial position of <OBJ\{id\}>.''}\\
\texttt{``How does <OBJ\{id\}> relate spatially to the surrounding objects?''}\\
\texttt{``Describe the location and spatial context of <OBJ\{id\}>.''}\\
\texttt{``Describe where <OBJ\{id\}> is relative to other objects in the scene.''}
\end{quote}
The expected answer is the original referring expression.

\paragraph{QA-I (Object Identification).}
Each sample presents a room-identifying spatial description and asks the model to identify a target object's class from four options.
To construct the question, we prompt Qwen3-14B~\cite{qwen3} to rewrite the target object's referring expression into an interrogative form that masks the target class name.
For example, ``the chair between the sofa and the bookshelf, next to the lamp'' becomes ``what is between the sofa and the bookshelf, next to the lamp?''---a transformation that is difficult to achieve reliably with rule-based templates due to the varied syntactic structures of multi-reference expressions.
The rewritten query is then concatenated with the room-identifying spatial description as a preamble, and three distractor classes are sampled from the same room; the four options are randomly shuffled to ensure uniform answer distribution across A/B/C/D.
Generated questions are validated to ensure the target class name does not appear in any form within the question text. Up to 3 questions are generated per target object. The full prompt template is shown in~\Cref{fig:qa1_prompt}.

\begin{figure}[h]
\centering
\begin{tcolorbox}[
  colback=gray!3,
  colframe=black!40,
  boxrule=0.4pt,
  width=0.95\linewidth,
  arc=2pt,
  left=8pt,
  right=8pt,
  top=6pt,
  bottom=6pt,
  title={\small Prompt Template for QA-I Question Generation},
  fonttitle=\bfseries\small
]
\small\ttfamily
Rewrite the following object description into a question that asks ``what'' or ``which object'' matches the description. Do NOT include the target class name anywhere in the question.\\[4pt]
Target description: \{target\_utterance\}\\
Target class (to be masked): \{target\_class\}\\[4pt]
Rules:\\
1) Output one line only, starting with ``Q:'' and ending with ``?''\\
2) Replace the target class with ``what'' or ``which object''\\
3) Keep all spatial constraints from the description\\
4) If you cannot comply, output: INVALID\\[4pt]
Example 1:\\
Input: the chair is behind the lamp and in front of the cabinet\\
Class: chair\\
Output: Q: What is behind the lamp and in front of the cabinet?\\[4pt]
Example 2:\\
Input: the lamp next to the desk that supports the monitor near the window\\
Class: lamp\\
Output: Q: What is next to the desk that supports the monitor near the window?
\end{tcolorbox}
\caption{Prompt template for generating QA-I questions. The target class name is masked in the output to form an identification question.}
\label{fig:qa1_prompt}
\end{figure}

\paragraph{QA-II (Counting).}
Each sample asks the model to count instances of a specified class within the room identified by the spatial description.
Only classes with 2--10 instances in the target room are included.
The spatial description is wrapped into a room phrase and combined with one of five templates, where \texttt{\{room\}} denotes the room phrase and \texttt{\{class\}} the pluralized category name:
\begin{quote}\small
\texttt{``In \{room\}, how many \{class\} are there?''}\\
\texttt{``In \{room\}, what is the total number of \{class\}?''}\\
\texttt{``How many \{class\} can be found in \{room\}?''}\\
\texttt{``How many \{class\} are there in \{room\}?''}\\
\texttt{``In \{room\}, how many \{class\} do you see?''}
\end{quote}
The expected answer is the integer count.

\paragraph{QA-III (Existence Verification).}
Each sample asks whether a specified class is present in the room identified by the spatial description.
For positive samples, the queried class must be present in the localized room but not mentioned in the spatial description, preventing trivial lookup.
For negative samples, the queried class is drawn exclusively from adjacent rooms and must not appear in the localized room, ensuring non-trivial distractors.
Positive and negative samples are balanced 1:1 per scene by truncating to the minority count.
Five prompt templates are used, with singular and plural variants selected based on the queried class, where \texttt{\{room\}} denotes the room phrase, \texttt{\{art\}} the appropriate article, and \texttt{\{class\}} the category name:
\begin{quote}\small
\texttt{``In \{room\}, is there \{art\} \{class\}?''}\\
\texttt{``Can you find \{art\} \{class\} in \{room\}?''}\\
\texttt{``Does \{room\} contain \{art\} \{class\}?''}\\
\texttt{``Is \{art\} \{class\} present in \{room\}?''}\\
\texttt{``In \{room\}, is \{art\} \{class\} present?''}
\end{quote}
The expected answer is ``Yes.'' or ``No.''.

\paragraph{QA-IV (Inter-Room Comparison).}
Each sample provides spatial descriptions for two distinct rooms and asks which contains more instances of a specified class.
Room pairs are drawn from all combinations within a scene, and the queried class must appear in both rooms with differing counts.
To ensure label balance, the assignment of rooms to options (A) and (B) is randomly swapped with probability 0.5 at generation time.
Three templates are used, where \texttt{\{room\_A\}} and \texttt{\{room\_B\}} denote the two spatial descriptions and \texttt{\{class\}} the pluralized category name:
\begin{quote}\small
\texttt{``Room A: "\{room\_A\}". Room B: "\{room\_B\}".}\\
\texttt{Which room has more \{class\}?''}\\[4pt]
\texttt{``Room A: "\{room\_A\}". Room B: "\{room\_B\}".}\\
\texttt{In which room can you find more \{class\}? ''}\\[4pt]
\texttt{``Room A: "\{room\_A\}". Room B: "\{room\_B\}".}\\
\texttt{Comparing both rooms, which has a greater number of \{class\}?''}
\end{quote}
The expected answer is ``A.'' or ``B.''.

\section{Implementation Details}
\label{app:appendix_implementation}

\subsection{Hyperparameters and Training}
\label{app:appendix_hyperparams}

We use AdamW with base learning rate $5\times10^{-6}$, betas $(0.9, 0.999)$, and weight decay $0.02$. The learning rate follows a cosine decay schedule with linear warmup over the first 10\% of epoch 1 and a min-LR multiplier of $0.01$. We use an aggressive max-grad-norm of $0.01$, empirically chosen to stabilize training of the bias and graph modules under the small base learning rate; values in the more common range $[1.0, 5.0]$ produced unstable loss curves on the hierarchical components in our preliminary experiments. The per-GPU batch size is 16 over 8 GPUs, giving a global effective batch size of 128. Both Stage~1 and Stage~2 train for 3 epochs in bf16 precision. We use random seed 42 for all reported numbers.

The LLM is adapted via LoRA~\cite{lora} with rank 16, alpha 16, and dropout 0.05, applied to all attention and MLP projections (\texttt{q\_proj}, \texttt{k\_proj}, \texttt{v\_proj}, \texttt{o\_proj}, \texttt{gate\_proj}, \texttt{up\_proj}, \texttt{down\_proj}). In addition to LoRA parameters, the input token embeddings and corresponding LM-head rows are unfrozen for the introduced object identifier tokens (\texttt{<OBJ\_001>}, $\ldots$, \texttt{<OBJ\_300>}) to enable end-to-end learning of grounding outputs.

All experiments are conducted on 8$\times$ NVIDIA H200 GPUs (141\,GB HBM3e per card), with steady-state training memory of approximately 80\,GB per GPU. The two-stage training takes approximately 12 wall-clock hours per backbone: Stage~1 (3 epochs on ScanNet-based tasks) takes $\sim$4 hours, and Stage~2 (3 epochs on CAIRN-MR plus 30\% ScanNet replay) takes $\sim$8 hours.

\subsection{Architectural Details}
\label{app:appendix_arch}

\paragraph{Object Token Block.}
Each object is represented as a 4-token block $[t^{id}, t^{3D}, t^{2D}, t^G]$. 
The 3D and 2D tokens, $t^{3D}$ and $t^{2D}$, are obtained by projecting 
pretrained features from Uni3D~\cite{uni3d} (applied to 
per-object point-cloud segments) and DINOv2~\cite{dinov2} 
(applied to multi-view rendered images of the object), respectively. 
All four tokens are projected to the LLM hidden dimension via 2-layer 
MLPs. The maximum number of objects per scene is capped at 300; 
task instances targeting objects beyond this cap are excluded from 
training and evaluation. Empirically, this cap retains over 96\% 
of task instances in CAIRN-MR and all task instances in the 
ScanNet-based benchmarks.

\paragraph{Message Passing for Graph Tokens.}
The graph token $t_i^G$ in the object token block encodes relational 
context beyond per-object geometry and appearance, complementing the 
3D and 2D tokens with information aggregated from neighboring objects. 
We instantiate the message-passing scheme outlined in \Cref{sec:tokenization} 
as follows.

For each object $\mathcal{O}_i$, we construct a room-local subgraph 
by connecting it to its $K_g$ nearest in-room neighbors under the 
edge set $\mathcal{E}_{OO}$ defined in \Cref{sec:attention}. 
Node features $\mathbf{h}_i^{(0)}$ are initialized by concatenating 
the per-object Uni3D point-cloud feature with a small geometric 
descriptor (center coordinates and bounding-box size). Edge features $\mathbf{e}_{ij}$ are constructed by concatenating a 
geometric component encoding relative position and Euclidean distance 
between $\mathcal{O}_i$ and $\mathcal{O}_j$, with a pretrained VL-SAT~\cite{wang2023vl} 
relation embedding capturing semantic spatial relations between the pair (e.g., above, next to). 
Both edge components are precomputed and held fixed during training.

We then perform $L_g$ layers of message passing:
\begin{equation}
    \mathbf{h}_i^{(\ell+1)} =
    \mathbf{h}_i^{(\ell)}
    +
    \Phi\!\left(
    \mathbf{h}_i^{(\ell)},\,
    \mathrm{AGG}\Big(
    \{\Psi(\mathbf{h}_j^{(\ell)},\, \mathbf{e}_{ij})
    \mid j \in \mathcal{N}(i)\}
    \Big)
    \right),
\end{equation}
where $\Psi$ computes a message from each neighbor's embedding and 
the edge attribute, $\mathrm{AGG}$ denotes mean aggregation, and 
$\Phi$ fuses the node's own embedding with the aggregated message. 
The final node embedding $\mathbf{h}_i^{(L_g)}$ is projected into 
the LLM hidden space by a 2-layer MLP to form the graph token $t_i^G$.

We use $L_g=2$ message-passing layers with mean aggregation, 
following the standard depth in GraphSAGE-style architectures~\cite{hamilton2017inductive}, 
where two layers suffice to propagate information across the immediate 
spatial neighborhood while avoiding the over-smoothing observed at 
greater depths.


\end{document}